\documentclass[10pt,twocolumn,letterpaper]{article}

\usepackage{iccv}              

\definecolor{iccvblue}{rgb}{0.21,0.49,0.74}
\usepackage[pagebackref,breaklinks,colorlinks,allcolors=iccvblue]{hyperref}
\usepackage{booktabs}
\usepackage{makecell}
\usepackage{colortbl}
\usepackage{xcolor}
\usepackage{tabularx}

\usepackage[utf8]{inputenc}
\usepackage[T1]{fontenc}
\usepackage{textcomp}

\def\paperID{7440} 
\def\confName{ICCV}
\def\confYear{2025}

\title{A Visual Leap in CLIP Compositionality Reasoning through Generation of Counterfactual Sets}

\author{
Zexi Jia\textsuperscript{1},
Chuanwei Huang\textsuperscript{2},
Hongyan Fei\textsuperscript{2},
Yeshuang Zhu\textsuperscript{1},
Zhiqiang Yuan\textsuperscript{1},
Ying Deng\textsuperscript{1},
\\
Jiapei Zhang\textsuperscript{1},
Jinchao Zhang\textsuperscript{1*},
Jie Zhou\textsuperscript{1}
\\
\textsuperscript{1}WeChat AI, Tencent Inc, China\\
\textsuperscript{2}Institute for Artificial Intelligence, Peking University
}

\begin{document}
\maketitle

\begin{abstract}
Vision-language models (VLMs) often struggle with compositional reasoning due to insufficient high-quality image-text data. To tackle this challenge, we propose a novel block-based diffusion approach that automatically generates counterfactual datasets without manual annotation. Our method utilizes large language models to identify entities and their spatial relationships. It then independently generates image blocks as "puzzle pieces" coherently arranged according to specified compositional rules. This process creates diverse, high-fidelity counterfactual image-text pairs with precisely controlled variations. In addition, we introduce a specialized loss function that differentiates inter-set from intra-set samples, enhancing training efficiency and reducing the need for negative samples. Experiments demonstrate that fine-tuning VLMs with our counterfactual datasets significantly improves visual reasoning performance. Our approach achieves state-of-the-art results across multiple benchmarks while using substantially less training data than existing methods.
\end{abstract}    

\section{Introduction}

\renewcommand{\thefootnote}{}

\footnote{* Corresponding Author. This work has been accepted by ICCV 2025.}

Vision-language models (VLMs) \cite{radford2021learning,liu2024visual,zhu2023minigpt,li2023blip,bai2023qwen,lin2024vila,xue2024xgen} have significantly advanced artificial intelligence applications such as image retrieval, visual question answering, and multimodal dialogue systems. Despite their progress, VLMs struggle with compositional reasoning—understanding the attributes, positions, and relationships among objects. For example, as shown in Figure 1(c), models like CLIP \cite{radford2021learning} often fail to distinguish attributes and spatial relations between people and doors or people and trains in complex scenes.

These limitations result in superficial interpretations and potential biases. When integrated with large language models to develop advanced multimodal dialogue systems \cite{chen2024internvl,chen2024far,mckinzie2024mm1,dong2024internlm}, the issues become more pronounced, causing errors in tasks like scene description, visual question answering, and visual reasoning. For instance, a model may misinterpret user queries about specific objects in an image, leading to incorrect responses.

We attribute these challenges to the lack of high-quality, detailed image-text pairs. Although large-scale datasets like LAION-400M \cite{schuhmann2021laion} and YFCC100M \cite{thomee2016yfcc100m} exist, annotation quality and alignment remain problematic. Previous works, such as ARO \cite{yuksekgonul2023and} and VL-Checklist \cite{zhao2022vl}, attempted to augment data by adjusting text order and replacing words but failed to produce more challenging and diverse image-text pairs.

Recent studies question the effectiveness of simple textual modifications for generating negative examples. Creating challenging negative captions, as shown in Figure 1(a), allows text encoders to achieve nearly perfect classification accuracy without relying on image information. However, Urbanek et al. \cite{urbanek2024picture} point out that improvements on ARO or VL-Checklist stem from identical construction patterns in both test and training sets; models fine-tuned on such data perform poorly on complex relational datasets like Winoground \cite{thrush2022winoground} and sDCI \cite{urbanek2024picture}. Therefore, developing more comprehensive methods for augmenting high-quality training data is crucial.

In this paper, we propose an innovative method to automatically augment dense counterfactual text-image sets without manual annotations. A counterfactual set comprises text-image pairs derived from an original image, differing in certain entity combinations. Mimicking jigsaw puzzles, we create variants by adjusting ``puzzle pieces'' within images—adding, removing, or modifying entities. By combining a small set of real image-text pairs with large language models (LLMs) and generative diffusion models, we flexibly generate substantial-high-quality counterfactual data.

Prior studies often avoid using generative models or rely only on local image editing techniques \cite{awal2025vismin,peng2024synthesize} to create challenging negative samples because generative models struggle to capture complex object relationships without precise guidance. To address this limitation, we combine a puzzle construction approach with the capabilities of LLMs, providing precise positional information to effectively guide the generative model.

We introduce a novel technique called \emph{block-based diffusion}, which integrates local descriptions of image elements with the global scene description within the diffusion process. Unlike previous diffusion-based image editing or collage methods, our approach leverages LLMs to parse textual descriptions, extracting key entities and determining their precise coordinates in the image.

We generate image blocks for these entities using a diffusion model, forming pieces of our ``puzzle'' and arrange them according to the specified compositional relationships. These blocks serve as local reference images and, along with their corresponding entity descriptions, are incorporated into the diffusion process through cross-attention as guidance. We implement a dynamic weighting scheme: during the early stages of diffusion, each block is generated independently; in later stages, greater weight is given to the overall scene description to ensure coherence. This method allows us to generate multiple sets of richly annotated counterfactual image-text pairs where the images closely align with the complex compositional relationships described in the text, exhibiting strong overall coherence (see Figure 1(b)).

Furthermore, recognizing the limitations of traditional contrastive loss fine-tuning, we design a new loss function specifically for the counterfactual dataset. This function independently evaluates similarities of positive and negative sample pairs within each set and computes similarities between sets using only representative samples. This reduces computational complexity and avoids reliance on a large number of negative samples. 

In summary, our main contributions include:

\begin{itemize} \item An automated pipeline for generating high-quality counterfactual image-text pairs with precise control over compositional attributes without manual annotation.

\item A block-based diffusion technique that treats objects as semantic "puzzle pieces," enabling accurate positioning while maintaining global scene coherence.

\item A novel paired loss function that transforms traditional contrastive learning into efficient sample-pair optimization, leveraging counterfactual set structure to reduce negative sample requirements and communication overhead.

\item Experimental validation showing our approach outperforms methods using significantly larger manually annotated datasets on compositional reasoning benchmarks. \end{itemize}

\section{Related Work}

\subsection{Vision-language Models}

Recent developments in Vision-Language Models (VLMs) \cite{radford2021learning,chen2023internvl,liu2024visual,zhu2023minigpt,xue2024xgen,gao2024mini,jia2025secret,huang2025semantic} demonstrate significant progress, largely due to pioneering works like CLIP \cite{radford2021learning}. CLIP introduces the use of large-scale image-text pairs and contrastive learning for joint pre-training of image and text encoders, laying a solid foundation for subsequent research. Building on this, BLIP \cite{li2022blip} advances the field by leveraging noisy web data to automatically generate captions and effectively addressing challenges in understanding and generation tasks. BLIP-2 \cite{li2023blip} improves efficiency in cross-modal information retrieval by using pre-trained image encoders and language models. Flava \cite{singh2022flava} enhances model performance by integrating multiple VLM objectives and auxiliary modality-specific losses. X-VLM \cite{zeng2021multi} achieves notable results by finely aligning text sub-components with corresponding image regions at multiple scales. Recently, LLaVA \cite{liu2024visual} stands out with its large-scale multimodal pre-training and unique unified architecture, demonstrating exceptional capabilities in handling a wide range of tasks.

Despite these achievements, VLMs still face challenges. One major issue is the mismatch between images and their text annotations. This mismatch not only involves content differences but also includes overly brief text descriptions that lack sufficient background information, affecting the scene understanding capabilities of VLMs. These challenges limit the effectiveness of VLMs in practical applications, especially in scenarios requiring precise understanding and generation of complex vision-language relationships \cite{fan2023advances,chen2023clip2scene}.

\subsection{Vision-language Compositionality Reasoning }

Vision-language compositionality reasoning (VLCR) \cite{gupta2023visual,zeng2024investigating,hsieh2024sugarcrepe} is a key technology that enables deep understanding and analysis of visual scenes combined with related text. This technology requires models to not only detect and recognize individual entities within images but also to understand textual descriptions and infer the relationships and attributes between these entities and the text.

VLCR focuses on understanding and processing the complex relationships within visual and linguistic information. Current research frequently utilizes pre-trained cross-modal models to improve noisy image-text pairs sourced from the internet. This approach leads to higher-quality descriptions and enhances visual-linguistic compositional reasoning. While one might expect vision-language models (VLMs) to naturally develop compositional reasoning skills through extensive training on large numbers of image-text pairs, most state-of-the-art VLMs exhibit characteristics more akin to "bag-of-words" models \cite{yuksekgonul2023and,hsieh2024sugarcrepe,yarom2024you}. These models are capable of aligning textual entities with visual elements in images but struggle to interpret the relationships or properties between them.

To address this issue, researchers have employed various strategies, such as integrating negative textual prompts into training sets \cite{yuksekgonul2023and,doveh2023teaching} or generating analogous descriptions using large language models (LLMs) \cite{doveh2024dense,urbanek2024picture}. However, these methods often generate overly simple data that fail to produce images matching the altered prompts. As a result, they do not enhance the model's ability to understand the complex relationships between images and text.

\section{Method}

\begin{figure*}[t]
\begin{center}
\includegraphics[width=1.0\linewidth]{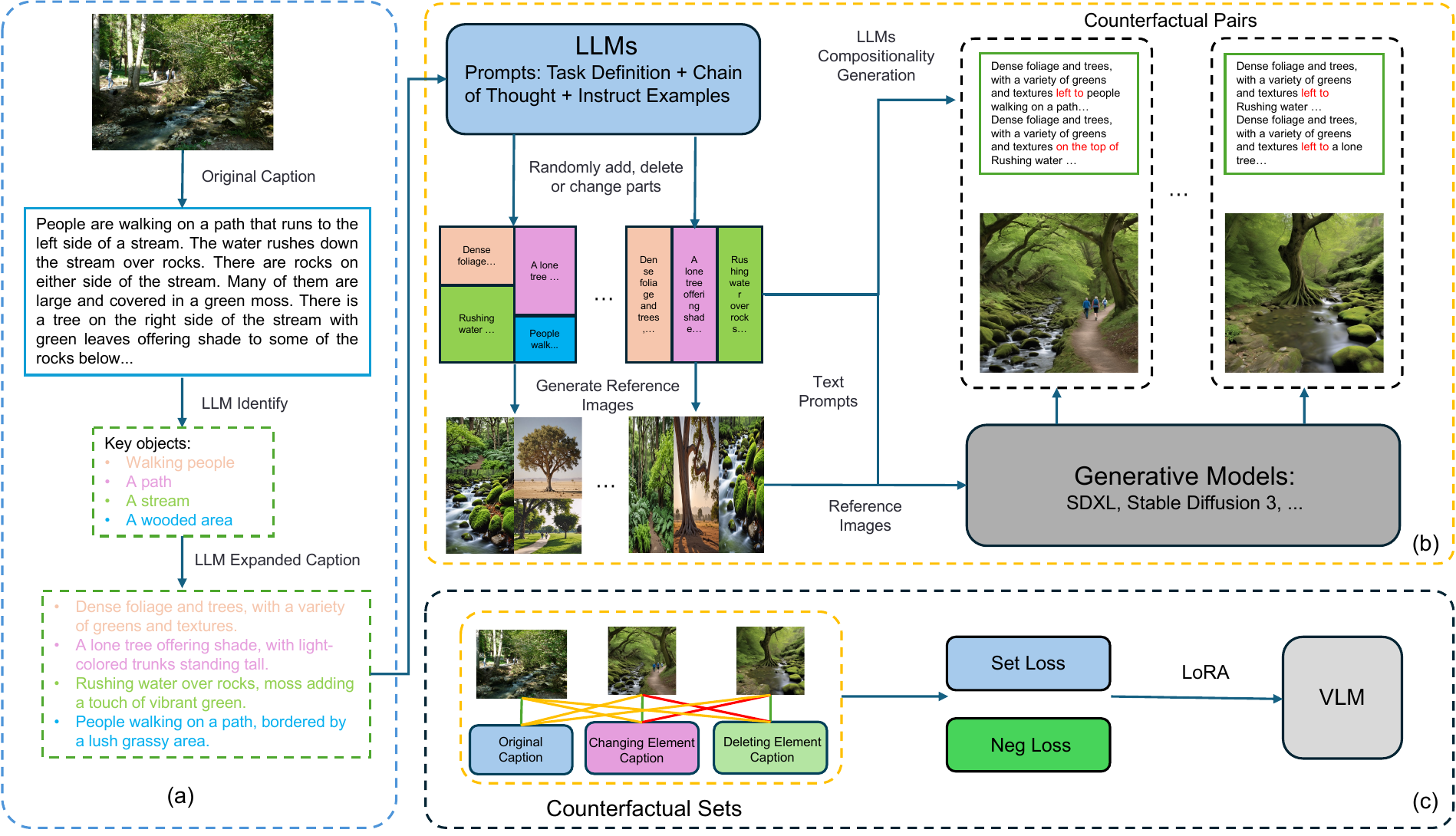}
\end{center}
   \caption{Generating image-text counterfactual sets involves several steps: (a) An LLM identifies and diversifies core entities from dense text descriptions. (b) Dense captions are processed through an LLM to obtain structured regional captions, which are used to generate individual images and a reference image. These regional captions are further processed to create a Compositionality Caption. (c) The reference image and Compositionality Caption are input into a diffusion model to generate images that match the Compositionality Caption, forming Counterfactual Pairs. Multiple Counterfactual Pairs are grouped into Counterfactual Sets, and specific loss functions are designed for these sets to efficiently fine-tune the text and image encoders of the CLIP model using LORA.}
\label{fig:first}
\end{figure*}

\subsection{Counterfactual Data Augmentation}

To enrich our training dataset, we incorporate synthetic counterfactual image-text pairs. Drawing inspiration from solving a jigsaw puzzle, we generate images that accurately reflect complex textual descriptions, precisely capturing entities' positions and attributes.

As shown in Figure 2(a), we utilize a Large Language Model (LLM) to parse textual descriptions, identifying key entities and enhancing description diversity by adjusting entity attributes such as color, type, state, or context. For instance, from \emph{a white dog on the left of a black cat}, the LLM identifies \emph{a white dog} and \emph{a black cat} and introduces variations like changing the black cat to a yellow one. The LLM also places each entity into specific scenes or actions, adding context like \emph{the white dog running on the grass}. These detailed descriptions guide image generation, enabling precise positioning and sizing of elements for seamless image composition.

Using these descriptions (Figure 2(b)), the LLM constructs regional descriptions specifying entities' positions and attributes. For example, \emph{the white dog running on the grass is to the left of the yellow cat dozing lazily}. These statements guide image generation, scaling, and combining images at predefined coordinates to form reference images. Using our block-based generation strategy, multiple diffusion processes are applied to image regions. These are adaptively blended according to regional semantic descriptions and reference images to produce outputs matching specified details precisely.

To ensure high consistency between generated images and counterfactual texts, we adopt several strategies: providing partially generated images as references to maintain entity attributes and positions; employing our proposed \textbf{block-based diffusion} where each block diffuses independently before integration; and using CLIP to filter out results with low text-image similarity. Through these measures, we construct a high-quality counterfactual dataset.

We decompose compositionality into three dimensions, attributes, positions, and relationships, to guide the LLM in detailed description expansion. To change attributes, we modify one object's attribute while keeping its position. To change positions, we adjust relative position descriptions without altering attributes. To change relationships, we limit the number of objects and use a Chain of Thought (CoT) to help the generative model analyze parts and relationships step by step.

As shown in Figure 2(c), we generate multiple counterfactual examples by adding, removing, or modifying elements and using LLMs to produce diverse descriptions. These examples form a counterfactual dataset that challenges the model in a complex environment, enhancing precision and reliability. Incorporating these counterfactual samples expands dataset diversity and complexity, enabling better understanding and generalization of compositional concepts.

\subsection{Block-based Diffusion Generation Strategy}

We propose a block-based diffusion strategy that enhances image synthesis by integrating holistic and local guidance within a unified framework. Our method decomposes the image generation process into multiple semantic regions, or \emph{blocks}, each corresponding to an entity or object in the scene. This approach addresses conflicts in overlapping regions and handles complex semantic relationships and attribute bindings.

We leverage Large Language Models (LLMs) to parse the input prompt and extract the overall scene description $T_{\text{global}}$, which is a comprehensive description of the scene; local region descriptions $\{ T_i \}$, providing detailed descriptions for each entity including attributes and spatial relationships; and positional coordinates $\{ P_i \}$, specifying precise spatial positions for each entity. For each entity, we generate a reference image $I_i$ using a text-to-image model conditioned on $T_i$, offering visual guidance for the entity's attributes and appearance.

Our image generation process is based on a denoising diffusion probabilistic model (DDPM). At each diffusion timestep $t$, we integrate global and local guidance into the denoising process by manipulating the cross-attention layers of the denoising network and adjusting their influence over time through dynamic weighting. The hidden state update at timestep $t$ is:

\begin{equation}
\mathbf{h}_t = \mathbf{h}_t + w_{\text{global}}(t) \cdot \text{Attn}_{\text{global}} + \sum_i w_{\text{local}}(t) \cdot M_i \cdot \text{Attn}_{i},
\end{equation}

where $\text{Attn}_{\text{global}}$ attends to the overall scene description $T_{\text{global}}$, and $\text{Attn}_{i}$ attends to the local description $T_i$ and local image guidance $I_i$ for entity $i$. The spatial mask $M_i$ restricts the influence to the predefined position $P_i$.

For local guidance, the attention $\text{Attn}_{i}$ combines text and image features:

\begin{equation}
\text{Attn}_{i} = \text{Softmax}\left( \frac{\mathbf{q} \left[ \mathbf{k}_{T_i}; \mathbf{k}_{I_i} \right]^\top }{ \sqrt{d} } \right) \left[ \mathbf{v}_{T_i}; \mathbf{v}_{I_i} \right],
\end{equation}

where $\mathbf{k}_{T_i}, \mathbf{v}_{T_i}$ are key and value embeddings from the local text description $T_i$, $\mathbf{k}_{I_i}, \mathbf{v}_{I_i}$ are from the local image guidance $I_i$, and $\left[ \, ; \, \right]$ denotes concatenation. Similarly, $\text{Attn}_{\text{global}}$ is computed from $T_{\text{global}}$.

The weighting functions to control the influence over time:

\begin{equation}
\begin{aligned}
w_{\text{local}}(t) &=
\begin{cases}
w_{\text{max}}, & t \leq t_{\text{th}} \\
w_{\text{max}} \left( 1 - \dfrac{t - t_{\text{th}}}{T - t_{\text{th}}} \right), & t > t_{\text{th}}
\end{cases} \\
w_{\text{global}}(t) &= w_{\text{max}} - w_{\text{local}}(t),
\end{aligned}
\end{equation}

where $T$ is the total number of diffusion timesteps, $t_{\text{th}}$ marks when we start decreasing the local guidance weight, and $w_{\text{max}}$ is typically set to $1$. This dynamic weighting allows local regions to be generated independently in early stages and integrates them into a coherent global image as diffusion progresses.

By applying spatial masks $M_i$, each local guidance influences only its corresponding region, enhancing detail fidelity and compositional accuracy. The local image guidance $I_i$ complements the textual descriptions, providing precise visual cues for each entity's appearance.

Through this strategy, we dynamically adjust the influence of local descriptions, local image guidance, and the overall description during the diffusion process. Our method effectively handles complex semantic relationships and attribute bindings, resulting in images that are globally consistent and rich in local details.

\subsection{Training Losses}

We consider dual-encoder VLMs as usually done in the literature, which processes a text-image pair ($x$, $y$) comprising two primary components: (i) an image encoder that transforms the image $x$ into a representation $\mathbf{z_x}$; (ii) a text encoder that converts the text $y$ into a representation $\mathbf{z_y}$. 
We use the normalized inner product $\mathcal{I}(x, y)$ as the cosine similarity to measure the similarity between $x$ and $y$: $\mathcal{I}(x, y)=\frac{\mathbf{z_x}^T \mathbf{z_y}}{\left\|\mathbf{z_x}\right\|\left\|\mathbf{z_y}\right\|}$.

\begin{figure}[h]
\begin{center}
\includegraphics[width=1.0\linewidth]{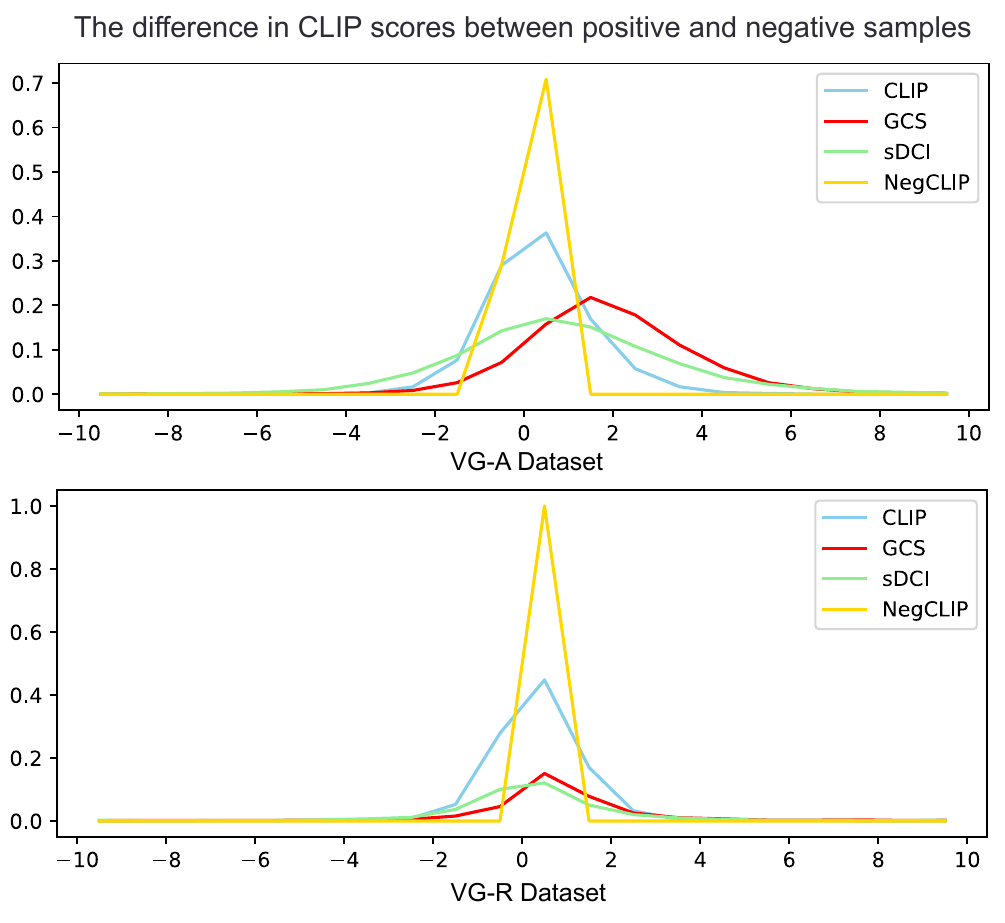}
\end{center}
   \caption{Compare CLIP-score differences between positive and negative descriptions on VG-A and VG-R datasets (Y-axis: proportion). Other fine-tuned models show limited improvement, while our method enlarges the gap.}
\label{fig:fig3}
\end{figure}

\begin{table*}[h!]
\centering
\caption{Performance: The accuracy of VLMs and fine-tuned CLIP on different datasets in the ARO and VL-Checklist benchmarks.}
\label{tab:exp1}
\begin{tabular*}{\textwidth}{@{\extracolsep{\fill}}lccccccc}
\toprule
\textbf{Model} 
& \multicolumn{4}{c}{\textbf{ARO}} & \multicolumn{3}{c}{\textbf{VL-Checklist}} \\
\cmidrule(lr){2-5} \cmidrule(lr){6-8}& Relation (VGR) & Attribute (VGA) & Order (COCO) & Order (Flickr) & Object & Attribute & Relation \\
\midrule
\multicolumn{8}{c}{\textbf{Vision-Language Models}} \\
\midrule
CLIP \cite{radford2021learning} & 59.9 & 62.9 & 47.9 & 59.0 & 81.1 & 67.6 & 61.9 \\
BLIP-2 \cite{li2023blip}  & 41.2 & 71.2 & 13.6 & 13.7 & 84.2 & 80.1 & 70.7 \\
\midrule
\multicolumn{8}{c}{\textbf{Fine-Tuned CLIP Models}} \\
\midrule
NegCLIP \cite{yuksekgonul2023and} & 81.1 & 70.9 & 85.5 & 90.9 & 81.3 & 72.2 & 63.5 \\
SVLC \cite{doveh2023teaching} & 80.6 & 73.0 & 84.7 & 91.7 & 85.0 & 72.0 & 68.9 \\
DAC$_{\text{LLM-10k}}$ \cite{doveh2023dense} & 61.5 & 63.9 & 46.7 & 60.1 & 66.9 & 57.4 & 57.0 \\
DAC$_{\text{LLM-3m}}$ \cite{doveh2023dense} & 81.3 & 73.9 & 94.4 & 95.4 & 87.3 & 77.2 & 86.4 \\
DAC$_{\text{SAM-3m}}$ \cite{doveh2023dense} & 77.1 & 70.5 & 91.2 & 93.8 & 88.5 & 75.8 & \textbf{89.7} \\
sDCI$_{\text{p1}}$ \cite{urbanek2024picture} & 76.2 & 67.5 & 88.4 & 91.1 & 80.7 & 68.7 & 70.1 \\
sDCI$_{\text{p1NL0}}$ \cite{urbanek2024picture} & 57.3 & 62.0 & 39.4 & 44.6 & 88.3 & 70.4 & 61.2 \\
\midrule
\textbf{GCS$_{\text{generate-10k}}$} & 82.2 & 67.7 & 93.9 & 92.5 & 83.4 & 71.9 & 74.5 \\
\textbf{GCS$_{\text{generate-300k}}$} & \textbf{85.7} & \textbf{73.4} & \textbf{94.6} & \textbf{95.6} & \textbf{89.1} & \textbf{81.4} & 85.6 \\
\bottomrule
\end{tabular*}
\end{table*}

\subsubsection{Previous Loss}

In previous research, VLMs and fine-tuning tasks have predominantly utilized the Contrastive Loss \cite{chen2020simple} as the primary optimization function. The form of the Contrastive loss is as follows:

\begin{equation}
\mathcal{L}_{cont}=-\sum_{i=1}\log \frac{e^{\left(\mathcal{I}(x_i, y_i)/ \tau\right)}}{\sum_{j=1} e^{\left(\mathcal{I}(x_i,y_j) / \tau\right)}},
\end{equation}
where $\tau$ is a learned temperature parameter. However, the contrastive loss only considers a single positive example and its effectiveness depends largely on the quality and difficulty of negative examples in the batch, leading to suboptimal similarity scores and data inefficiency.

\subsubsection{Proposed Loss Function}

We propose an efficient loss function for fine-tuning the CLIP model on our counterfactual sets. Unlike the Sigmoid Loss from \cite{zhai2023sigmoid}, which increases computational demands by assessing all image-text pairs, our loss divides the computation into intra-set and inter-set components.

For the \textbf{intra-set} loss $\mathcal{L}_{\text{intra}}$, we compute the similarity between positives and negatives within each counterfactual set:
\begin{equation}
\mathcal{L}_{\text{intra}} = - \sum_{i=1}^{m} \sum_{j=1}^{m} \log \frac{1}{1 + e^{ l_{ij} ( -\tau \mathcal{I}(x_i, y_j) + b ) }},
\end{equation}
where $l_{ij}$ is the label for an image-text pair ($1$ for a correct match, $-1$ otherwise), $b$ is a learnable bias term, and $m$ is the number of elements in a counterfactual set.

For the \textbf{inter-set} loss $\mathcal{L}_{\text{inter}}$, we use the real image-text pair $(x_i^0, y_i^0)$ from each set as a reference and consider real pairs from other sets as negatives:
\begin{equation}
\mathcal{L}_{\text{inter}} = - \sum_{i=1}^{n} \sum_{\substack{j=1 \\ j \neq i}}^{n} \log \frac{1}{1 + e^{ \tau \mathcal{I}(x_i^0, y_j^0) - b }},
\end{equation}
where $n$ is the number of counterfactual sets. The total loss combines both components:
\begin{equation}
\mathcal{L}_{\text{sets}} = \mathcal{L}_{\text{inter}} + \sum_{i=1}^{n} \mathcal{L}_{\text{intra}}^{(i)}.
\end{equation}
This loss function includes gradients from positive and negative examples within counterfactual sets and from representative samples across different sets.

\textbf{Negative-text loss}. Following \cite{doveh2023teaching}, we introduce a negative-text loss $\mathcal{L}_{\text{neg}}$ to help the model distinguish descriptions with inverted word orders:
\begin{equation}
\mathcal{L}_{\text{neg}} = - \sum_i \log \left( \frac{ e^{ \mathcal{I}(x_i, y_i) / \tau } }{ e^{ \mathcal{I}(x_i, y_i) / \tau } + e^{ \mathcal{I}(x_i^{\text{neg}}, y_i) / \tau } } \right),
\end{equation}
where $x_i^{\text{neg}}$ is the negative text generated by permuting the word order.

\textbf{Final loss}. The overall fine-tuning loss is:
\begin{equation}
\mathcal{L} = \mathcal{L}_{\text{sets}} + \mathcal{L}_{\text{neg}}.
\end{equation}
By optimizing this loss function, we increase the CLIP-score gap between correct and perturbed descriptions, enabling our fine-tuned model to better distinguish positive from negative examples while maintaining robustness and generalization (see Figure~3).

\section{Experimental Results}

\begin{table*}[h!]
\centering
\caption{Performance: The accuracy of CLIP-based fine-tuned models on complex relation datasets. SCM measures matching accuracy for cropped regions, Neg evaluates the ability to distinguish real from fake descriptions.}
\label{tab:exp2}
\begin{tabular*}{\textwidth}{@{\extracolsep{\fill}}lccccccc}
\toprule
\textbf{Model}  & \multicolumn{4}{c}{\textbf{sDCI}} & \multicolumn{3}{c}{\textbf{Winoground}} \\
\cmidrule(lr){2-5} \cmidrule(lr){6-8}
 & SCM$@1$ & Neg$@1$ & SCM$@5$ & Neg$@5$ & Text & Image & Group \\
\midrule
CLIP \cite{radford2021learning} & 40.1 & 62.9 & 11.2 & 59.0 & 31.3 & 11.3 & 9.0 \\
NegCLIP \cite{yuksekgonul2023and} & 43.2 & 56.0 & 13.2 & 34.8 & 29.3 & 10.5 & 8.0 \\
DAC$_{\text{LLM-3m}}$ \cite{doveh2023dense} & 37.4 & 81.7 & 8.1 & 37.2 & 25.3 & 13.5 & 7.3 \\
DAC$_{\text{SAM-3m}}$ \cite{doveh2023dense} & 37.9 & 84.7 & 6.7 & 39.8 & 28.3 & 14.5 & 8.5 \\
sDCI$_{\text{p1}}$ \cite{urbanek2024picture} & 45.2 & 88.2 & 12.5 & 77.7 & 20.8 & 10.3 & 5.3 \\
sDCI$_{\text{p1NL0}}$ \cite{urbanek2024picture} & 47.6 & 69.8 & 15.3 & 30.4 & 30.5 & 10.8 & 8.3 \\
\midrule
\textbf{GCS$_{\text{generate-10k}}$} & 36.7 & 82.2 & 9.4 & 78.2 & 23.8 & 17.3 & 8.3 \\
\textbf{GCS$_{\text{generate-300k}}$} & \textbf{47.9} & \textbf{90.2} & \textbf{15.6} & \textbf{79.8} & \textbf{32.8} & \textbf{19.5} & \textbf{10.0} \\
\bottomrule
\end{tabular*}
\end{table*}

\subsection{Datasets and Metrics}

\textbf{Training:} Following the DAC framework \cite{doveh2023dense}, we systematically augment the original brief COCO captions \cite{chen2015microsoft} (comprising over 200,000 images with 5+ captions each) using large language models. This process generates highly detailed descriptions that explicitly encode spatial relationships and object attributes. These augmented descriptions then serve as the basis for constructing our counterfactual image sets.

\textbf{Evaluation:} Our method is evaluated on four reasoning benchmarks and one classification benchmark:

VL-Checklist \cite{zhao2022vl} contains four sub-datasets with positive captions and negative captions created by altering one word to change the meaning.

ARO \cite{yuksekgonul2023and} comprises four sub-datasets (VG-R, VG-A, COCO, Flickr30K) with negative sentences created by reordering words to alter semantics, testing relationship understanding, attribute recognition, and word order sensitivity.

Winoground \cite{thrush2022winoground} provides expert-curated image-caption pairs requiring precise word order understanding, with two images matched to two captions using identical words in different arrangements.

sDCI \cite{urbanek2024picture} features dense annotations for full images and cropped regions with both human and LLM assistance. It introduces two metrics: SCM (accuracy of matching cropped regions to titles) and Neg (ability to distinguish real/fake titles).

Elevater \cite{li2022elevater} includes 20 image classification datasets used for zero-shot evaluation to verify retained general performance after fine-tuning.

\textbf{Metrics:} We evaluate using match accuracy, with sDCI requiring all description-image pairs to be correctly matched simultaneously.

\subsection{Implementation Details}
We adopt CLIP ViT/B-32 with batch size 100, learning rate 1e-5, and weight decay 0.1, fine-tune for 10 epochs using LoRA~\cite{hu2021lora} ($r=\alpha=32$). We utilize GPT-4o \cite{achiam2023gpt} as our LLM and employ SD-XL \cite{podell2023sdxl}, Stable Diffusion 3 \cite{esser2024scaling}, and PixArt-$\alpha$ \cite{chen2023pixart} as text-to-image models. All experiments run on 4 Nvidia V100 GPUs, with reported metrics representing averages from three runs using publicly available code and models from each work.

Our counterfactual dataset is equally distributed with 25\% each for modifications, additions, deletions, and regenerations with unchanged subjects. We balance this with real-data stitched images we create from the COCO dataset \cite{chen2015microsoft} using absolute coordinate relationships from the LLM and composite descriptions based on position coordinates and image descriptions. Our ablation studies analyze the impact of these stitched images.

\subsection{Performance}

\begin{table*}[h!]
\centering
\caption{Ablation study on the impact of loss functions, data, and set size.}
\label{tab:ab}
\setlength{\tabcolsep}{1.0mm}
\renewcommand{\arraystretch}{1.0}
\newcolumntype{Y}{>{\centering\arraybackslash}X}

\begin{tabularx}{\textwidth}{c|YYYYY|YYYY|YYY}
\toprule
     & \multicolumn{5}{c|}{\textbf{Ablation}} & \multicolumn{4}{c|}{\textbf{ARO}} & \multicolumn{3}{c}{\textbf{Winoground}} \\ 
\midrule
     & $L_{sets}$ & Stitched & Generated & $ L_{neg}$ & Set size & VGR & VGA & COCO & Flickr & Text & Image & Group \\ 
\midrule
CLIP & & & & & & 59.9 & 62.9 & 47.9 & 59.0 & 31.3 & 11.3 & 9.0 \\ 
\midrule
     &$L_{cont}$ & & & & 10& 62.8 & 60.1 & 70.2 & 61.6 & 14.5 & 14.3 & 5.0 \\ 
     & \scalebox{1.1}{$\checkmark$} & & & & 10& 82.3 & 64.5 & 76.1 & 72.3 & 15.0 & 14.8 & 5.3 \\ 
     & \scalebox{1.1}{$\checkmark$} & \scalebox{1.1}{$\checkmark$} & & & 10& 83.0 & 69.8 & 90.8 & 87.8 & 19.3 & 15.3 & 7.8 \\ 
Fine-tuning & \scalebox{1.1}{$\checkmark$} & \scalebox{1.1}{$\checkmark$} & \scalebox{1.1}{$\checkmark$} & & 10& 83.1 & 70.4 & 95.3 & 93.0 & 21.5 & 17.5 & 8.3 \\ 
      & \scalebox{1.1}{$\checkmark$} & \scalebox{1.1}{$\checkmark$} & \scalebox{1.1}{$\checkmark$} & \scalebox{1.1}{$\checkmark$} & 10 & 85.6 & 73.2 & \textbf{94.7} & 95.5 & 32.6 & 19.2 & \textbf{10.1} \\ 
   & \scalebox{1.1}{$\checkmark$} & \scalebox{1.1}{$\checkmark$} & \scalebox{1.1}{$\checkmark$} & \scalebox{1.1}{$\checkmark$} & 5 & 84.9 & 73.0 & 94.3 & 95.1 & 32.6 & 18.1 & 9.2 \\ 
     & \scalebox{1.1}{$\checkmark$} & \scalebox{1.1}{$\checkmark$} & \scalebox{1.1}{$\checkmark$} & \scalebox{1.1}{$\checkmark$} & 20 & \textbf{85.7} & \textbf{73.4} & 94.6 & \textbf{95.6} & \textbf{32.8} & \textbf{19.5} & 10.0 \\ 
\bottomrule
\end{tabularx}
\end{table*}

\begin{table}[h!]
\centering
\caption{Accuracy (\%) Across Guidance Types}
\label{tab:acc_compact}
\begin{tabular}{@{}p{1.2cm}@{\hspace{0.9em}}c@{\hspace{0.9em}}c@{\hspace{0.9em}}c@{\hspace{0.9em}}c@{}}
\toprule
\textbf{Feature} & \textbf{Baseline} & \textbf{Local img} & \textbf{Local text} & \textbf{Combined} \\
\midrule
Attribute     & 84.5 & 90.2 & 93.8 & \textbf{99.6} \\
Position      & 81.7 & 88.5 & 91.4 & \textbf{98.8} \\
Relation  & 76.9 & 85.1 & 89.3 & \textbf{95.2} \\
\bottomrule
\end{tabular}
\end{table}

\begin{table}[h!]
\centering
\caption{Performance comparison on Elevater benchmark.}
\label{tab:ab_performance}
\begin{tabular*}{\columnwidth}{@{\extracolsep{\fill}}lcccc}
\toprule
\textbf{Model} & \textbf{5-shot} & \textbf{10-shot} & \textbf{20-shot} & \textbf{All-shot} \\
\midrule
CLIP \cite{radford2021learning} & 66.19 & 69.58 & 71.90 & 78.96\\
DAC \cite{doveh2023dense}      & 64.92 & 69.20 & 72.98 & 77.44\\
GCS                            & 65.62 & 69.27 & 72.45 & 78.50 \\
\bottomrule
\end{tabular*}
\end{table}

Our method outperforms various VLMs on multiple benchmarks without using specially formatted training data. As Table~\ref{tab:exp1} shows, our counterfactual set generation (GCS) fine-tuning surpasses all baselines across ARO dataset subsets and maintains advantages on VL-Checklist, even against models trained with 10× more data.

For complex visual reasoning requiring spatial understanding, as shown in Table~\ref{tab:exp2}, GCS demonstrates superior performance. On the sDCI dataset, GCS achieves higher accuracy in global pairing (47.9\%), true/false description discrimination (90.2\%), and multi-image matching tasks (15.6\% and 79.8\% respectively). GCS also outperforms other CLIP-based models on the challenging Winoground dataset.

Notably, GCS excels at matching cropped image parts with descriptions despite no specific training for such tasks, surpassing even models fine-tuned with large region-matching datasets. In contrast, models using text perturbation methods like NegCLIP and DAC underperform the original CLIP baseline on complex visual reasoning.

These results demonstrate that even limited counterfactual sets effectively enhance VLMs' understanding of compositional relationships, with performance scaling with counterfactual set size.

\subsection{Quality Evaluation of Counterfactual Sets} 

We randomly sample 3000 instances from our dataset (1000 each for attributes, positions, and relationships) to evaluate our generated counterfactual data. Our analysis shows that our proposed combined approach using local image and local text guidance significantly outperforms the baseline method that relies solely on global descriptions. As shown in Table~\ref{tab:acc_compact}, the combined approach achieves accuracy rates of 99.6\% for attribute modification, 98.8\% for position modification, and 95.2\% for relationship modification, representing substantial improvements over the baseline. For relationship modifications, we create a whitelist of approximately 800 common interactive relationships to handle cases where entity swapping doesn't alter semantics (e.g., ``\emph{Jack and Mary are spouses}''), further enhancing our counterfactual dataset quality.

\subsection{Ablation Study}

In Table~\ref{tab:ab}, we conduct extensive ablation studies to evaluate the effects of our new loss function, merged and generated data, and the size of counterfactual datasets on enhancing the compositional understanding capabilities of VLMs. On the manually annotated Winoground dataset, we find that our proposed data and loss innovations improve model reasoning performance, both when used alone and in combination. In the ARO dataset, which is constructed by shuffling word orders, performance gains are primarily due to the new loss function on the VG-R and COCO sub-datasets. Conversely, on the VG-A and Flickr sub-datasets, improvements are mainly attributed to data generation. We also observe that increasing the number of elements in the counterfactual dataset slightly enhances model performance across all datasets. All experiments are conducted with the same number of training epochs and starting from the same pre-trained CLIP weights.

Furthermore, we compare the efficiency of our proposed loss function. Experiments under the same configuration demonstrate that our loss function reduces training time by an average of 13.6\% compared to the comparative loss function and by 16.2\% compared to the sigmoid loss when fine-tuning on 4 V100 GPUs. 

We also use the Elevater image classification benchmark \cite{li2022elevater} to evaluate the performance of our fine-tuned CLIP model when linearly transferred to downstream tasks. This evaluation is crucial as it indirectly demonstrates the practicality of our improved CLIP in various current downstream applications. Table~\ref{tab:ab_performance} shows the evaluation results of Elevater linear probing, where the performance of our GCS model matches that of the original CLIP model. This indicates that our data enhancement strategy does not impair the representational power of the original CLIP model.

To validate the broad applicability of our strategy, we fine-tuned BLIP-2 (dual-encoder) and MiniGPT-4 (multimodal LLM) using the same counterfactual dataset. As shown in Table~\ref{tab:cod_accuracy}, both architectures achieved performance gains, confirming that our method effectively enhances reasoning capabilities across model types.

\begin{table}[h!]
\centering
\caption{Experimental results on more VLMs.}
\label{tab:cod_accuracy}
\begin{tabular}{@{}lcccc@{}}
\toprule
\textbf{Model} & \textbf{ARO-A} & \textbf{ARO-R} & \textbf{Winoground} & \textbf{VL} \\

\midrule
BLIP-2 \cite{liu2024visual} & 71.2 & 41.2 & 28.3 & 78.3 \\
GCS-10k & 73.8 & 53.4 & 29.0 & 81.5 \\
GCS-300k & \textbf{76.3} & \textbf{74.9} & \textbf{30.9} & \textbf{86.4} \\

\midrule
MiniGPT \cite{zhu2023minigpt} & 55.7 & 46.9 & 8.3 & 78.7 \\
GCS-10k & 63.9 & 53.1 & 11.0 & 79.9 \\
GCS-300k & \textbf{71.0} & \textbf{60.2} & \textbf{15.3} & \textbf{82.1} \\
\bottomrule
\end{tabular}
\end{table}

\vspace{-5pt}
\section{Conclusion}
We present a block-based diffusion approach that generates counterfactual image-text pairs by treating objects as semantic puzzle pieces. Our method automatically synthesizes training samples with precise compositional control without manual annotation. By combining LLMs with our specialized diffusion model, we create structured counterfactual sets that capture fine-grained visual differences. Our efficient paired loss function eliminates extensive negative sampling requirements, reducing computational costs. Experiments show our synthetic data enhances model performance beyond larger manually annotated datasets, particularly for complex spatial reasoning tasks.

{
    \small
    \bibliographystyle{ieeenat_fullname}
    \bibliography{main}

\begin{thebibliography}{44}
\providecommand{\natexlab}[1]{#1}
\providecommand{\url}[1]{\texttt{#1}}
\expandafter\ifx\csname urlstyle\endcsname\relax
  \providecommand{\doi}[1]{doi: #1}\else
  \providecommand{\doi}{doi: \begingroup \urlstyle{rm}\Url}\fi

\bibitem[Achiam et~al.(2023)Achiam, Adler, Agarwal, Ahmad, Akkaya, Aleman, Almeida, Altenschmidt, Altman, Anadkat, et~al.]{achiam2023gpt}
Josh Achiam, Steven Adler, Sandhini Agarwal, Lama Ahmad, Ilge Akkaya, Florencia~Leoni Aleman, Diogo Almeida, Janko Altenschmidt, Sam Altman, Shyamal Anadkat, et~al.
\newblock Gpt-4 technical report.
\newblock \emph{arXiv preprint arXiv:2303.08774}, 2023.

\bibitem[Awal et~al.(2025)Awal, Ahmadi, Zhang, and Agrawal]{awal2025vismin}
Rabiul Awal, Saba Ahmadi, Le Zhang, and Aishwarya Agrawal.
\newblock Vismin: Visual minimal-change understanding.
\newblock \emph{Advances in Neural Information Processing Systems}, 37:\penalty0 107795--107829, 2025.

\bibitem[Bai et~al.(2023)Bai, Bai, Yang, Wang, Tan, Wang, Lin, Zhou, and Zhou]{bai2023qwen}
Jinze Bai, Shuai Bai, Shusheng Yang, Shijie Wang, Sinan Tan, Peng Wang, Junyang Lin, Chang Zhou, and Jingren Zhou.
\newblock Qwen-vl: A frontier large vision-language model with versatile abilities.
\newblock \emph{arXiv preprint arXiv:2308.12966}, 2023.

\bibitem[Chen et~al.(2023{\natexlab{a}})Chen, Yu, Ge, Yao, Xie, Wu, Wang, Kwok, Luo, Lu, et~al.]{chen2023pixart}
Junsong Chen, Jincheng Yu, Chongjian Ge, Lewei Yao, Enze Xie, Yue Wu, Zhongdao Wang, James Kwok, Ping Luo, Huchuan Lu, et~al.
\newblock Pixart-$alpha$: Fast training of diffusion transformer for photorealistic text-to-image synthesis.
\newblock \emph{arXiv preprint arXiv:2310.00426}, 2023{\natexlab{a}}.

\bibitem[Chen et~al.(2023{\natexlab{b}})Chen, Liu, Kong, Zhu, Ma, Li, Hou, Qiao, and Wang]{chen2023clip2scene}
Runnan Chen, Youquan Liu, Lingdong Kong, Xinge Zhu, Yuexin Ma, Yikang Li, Yuenan Hou, Yu Qiao, and Wenping Wang.
\newblock Clip2scene: Towards label-efficient 3d scene understanding by clip.
\newblock In \emph{Proceedings of the IEEE/CVF Conference on Computer Vision and Pattern Recognition}, pages 7020--7030, 2023{\natexlab{b}}.

\bibitem[Chen et~al.(2020)Chen, Kornblith, Norouzi, and Hinton]{chen2020simple}
Ting Chen, Simon Kornblith, Mohammad Norouzi, and Geoffrey Hinton.
\newblock A simple framework for contrastive learning of visual representations.
\newblock In \emph{International conference on machine learning}, pages 1597--1607. PMLR, 2020.

\bibitem[Chen et~al.(2015)Chen, Fang, Lin, Vedantam, Gupta, Doll{\'a}r, and Zitnick]{chen2015microsoft}
Xinlei Chen, Hao Fang, Tsung-Yi Lin, Ramakrishna Vedantam, Saurabh Gupta, Piotr Doll{\'a}r, and C~Lawrence Zitnick.
\newblock Microsoft coco captions: Data collection and evaluation server.
\newblock \emph{arXiv preprint arXiv:1504.00325}, 2015.

\bibitem[Chen et~al.(2023{\natexlab{c}})Chen, Wu, Wang, Su, Chen, Xing, Zhong, Zhang, Zhu, Lu, Li, Luo, Lu, Qiao, and Dai]{chen2023internvl}
Zhe Chen, Jiannan Wu, Wenhai Wang, Weijie Su, Guo Chen, Sen Xing, Muyan Zhong, Qinglong Zhang, Xizhou Zhu, Lewei Lu, Bin Li, Ping Luo, Tong Lu, Yu Qiao, and Jifeng Dai.
\newblock Internvl: Scaling up vision foundation models and aligning for generic visual-linguistic tasks.
\newblock \emph{arXiv preprint arXiv:2312.14238}, 2023{\natexlab{c}}.

\bibitem[Chen et~al.(2024{\natexlab{a}})Chen, Wang, Tian, Ye, Gao, Cui, Tong, Hu, Luo, Ma, et~al.]{chen2024far}
Zhe Chen, Weiyun Wang, Hao Tian, Shenglong Ye, Zhangwei Gao, Erfei Cui, Wenwen Tong, Kongzhi Hu, Jiapeng Luo, Zheng Ma, et~al.
\newblock How far are we to gpt-4v? closing the gap to commercial multimodal models with open-source suites.
\newblock \emph{arXiv preprint arXiv:2404.16821}, 2024{\natexlab{a}}.

\bibitem[Chen et~al.(2024{\natexlab{b}})Chen, Wu, Wang, Su, Chen, Xing, Zhong, Zhang, Zhu, Lu, et~al.]{chen2024internvl}
Zhe Chen, Jiannan Wu, Wenhai Wang, Weijie Su, Guo Chen, Sen Xing, Muyan Zhong, Qinglong Zhang, Xizhou Zhu, Lewei Lu, et~al.
\newblock Internvl: Scaling up vision foundation models and aligning for generic visual-linguistic tasks.
\newblock In \emph{Proceedings of the IEEE/CVF Conference on Computer Vision and Pattern Recognition}, pages 24185--24198, 2024{\natexlab{b}}.

\bibitem[Dong et~al.(2024)Dong, Zhang, Zang, Cao, Wang, Ouyang, Wei, Zhang, Duan, Cao, et~al.]{dong2024internlm}
Xiaoyi Dong, Pan Zhang, Yuhang Zang, Yuhang Cao, Bin Wang, Linke Ouyang, Xilin Wei, Songyang Zhang, Haodong Duan, Maosong Cao, et~al.
\newblock Internlm-xcomposer2: Mastering free-form text-image composition and comprehension in vision-language large model.
\newblock \emph{arXiv preprint arXiv:2401.16420}, 2024.

\bibitem[Doveh et~al.(2023{\natexlab{a}})Doveh, Arbelle, Harary, Herzig, Kim, Cascante-Bonilla, Alfassy, Panda, Giryes, Feris, et~al.]{doveh2023dense}
Sivan Doveh, Assaf Arbelle, Sivan Harary, Roei Herzig, Donghyun Kim, Paola Cascante-Bonilla, Amit Alfassy, Rameswar Panda, Raja Giryes, Rogerio Feris, et~al.
\newblock Dense and aligned captions (dac) promote compositional reasoning in vl models.
\newblock \emph{Advances in Neural Information Processing Systems}, 36:\penalty0 76137--76150, 2023{\natexlab{a}}.

\bibitem[Doveh et~al.(2023{\natexlab{b}})Doveh, Arbelle, Harary, Schwartz, Herzig, Giryes, Feris, Panda, Ullman, and Karlinsky]{doveh2023teaching}
Sivan Doveh, Assaf Arbelle, Sivan Harary, Eli Schwartz, Roei Herzig, Raja Giryes, Rogerio Feris, Rameswar Panda, Shimon Ullman, and Leonid Karlinsky.
\newblock Teaching structured vision \& language concepts to vision \& language models.
\newblock In \emph{Proceedings of the IEEE/CVF Conference on Computer Vision and Pattern Recognition}, pages 2657--2668, 2023{\natexlab{b}}.

\bibitem[Doveh et~al.(2024)Doveh, Arbelle, Harary, Herzig, Kim, Cascante-Bonilla, Alfassy, Panda, Giryes, Feris, et~al.]{doveh2024dense}
Sivan Doveh, Assaf Arbelle, Sivan Harary, Roei Herzig, Donghyun Kim, Paola Cascante-Bonilla, Amit Alfassy, Rameswar Panda, Raja Giryes, Rogerio Feris, et~al.
\newblock Dense and aligned captions (dac) promote compositional reasoning in vl models.
\newblock \emph{Advances in Neural Information Processing Systems}, 36, 2024.

\bibitem[Esser et~al.(2024)Esser, Kulal, Blattmann, Entezari, M{\"u}ller, Saini, Levi, Lorenz, Sauer, Boesel, et~al.]{esser2024scaling}
Patrick Esser, Sumith Kulal, Andreas Blattmann, Rahim Entezari, Jonas M{\"u}ller, Harry Saini, Yam Levi, Dominik Lorenz, Axel Sauer, Frederic Boesel, et~al.
\newblock Scaling rectified flow transformers for high-resolution image synthesis.
\newblock In \emph{Forty-first International Conference on Machine Learning}, 2024.

\bibitem[Fan et~al.(2023)Fan, Ji, Xu, Cheng, Sakaridis, and Van~Gool]{fan2023advances}
Deng-Ping Fan, Ge-Peng Ji, Peng Xu, Ming-Ming Cheng, Christos Sakaridis, and Luc Van~Gool.
\newblock Advances in deep concealed scene understanding.
\newblock \emph{Visual Intelligence}, 1\penalty0 (1):\penalty0 16, 2023.

\bibitem[Gao et~al.(2024)Gao, Chen, Cui, Ren, Wang, Zhu, Tian, Ye, He, Zhu, et~al.]{gao2024mini}
Zhangwei Gao, Zhe Chen, Erfei Cui, Yiming Ren, Weiyun Wang, Jinguo Zhu, Hao Tian, Shenglong Ye, Junjun He, Xizhou Zhu, et~al.
\newblock Mini-internvl: A flexible-transfer pocket multimodal model with 5\% parameters and 90\% performance.
\newblock \emph{arXiv preprint arXiv:2410.16261}, 2024.

\bibitem[Gupta and Kembhavi(2023)]{gupta2023visual}
Tanmay Gupta and Aniruddha Kembhavi.
\newblock Visual programming: Compositional visual reasoning without training.
\newblock In \emph{Proceedings of the IEEE/CVF Conference on Computer Vision and Pattern Recognition}, pages 14953--14962, 2023.

\bibitem[Hsieh et~al.(2024)Hsieh, Zhang, Ma, Kembhavi, and Krishna]{hsieh2024sugarcrepe}
Cheng-Yu Hsieh, Jieyu Zhang, Zixian Ma, Aniruddha Kembhavi, and Ranjay Krishna.
\newblock Sugarcrepe: Fixing hackable benchmarks for vision-language compositionality.
\newblock \emph{Advances in neural information processing systems}, 36, 2024.

\bibitem[Hu et~al.(2021)Hu, Shen, Wallis, Allen-Zhu, Li, Wang, Wang, and Chen]{hu2021lora}
Edward~J Hu, Yelong Shen, Phillip Wallis, Zeyuan Allen-Zhu, Yuanzhi Li, Shean Wang, Lu Wang, and Weizhu Chen.
\newblock Lora: Low-rank adaptation of large language models.
\newblock \emph{arXiv preprint arXiv:2106.09685}, 2021.

\bibitem[Huang et~al.(2025)Huang, Jia, Fei, Zhu, Yuan, Zhang, and Zhou]{huang2025semantic}
Chuanwei Huang, Zexi Jia, Hongyan Fei, Yeshuang Zhu, Zhiqiang Yuan, Jinchao Zhang, and Jie Zhou.
\newblock Semantic to structure: Learning structural representations for infringement detection.
\newblock \emph{arXiv preprint arXiv:2502.07323}, 2025.

\bibitem[Jia et~al.(2025)Jia, Huang, Zhu, Fei, Duan, Yuan, Deng, Zhang, Zhang, and Zhou]{jia2025secret}
Zexi Jia, Chuanwei Huang, Yeshuang Zhu, Hongyan Fei, Xiaoyue Duan, Zhiqiang Yuan, Ying Deng, Jiapei Zhang, Jinchao Zhang, and Jie Zhou.
\newblock Secret lies in color: Enhancing ai-generated images detection with color distribution analysis.
\newblock In \emph{Proceedings of the Computer Vision and Pattern Recognition Conference}, pages 13445--13454, 2025.

\bibitem[Li et~al.(2022{\natexlab{a}})Li, Liu, Li, Zhang, Aneja, Yang, Jin, Hu, Liu, Lee, et~al.]{li2022elevater}
Chunyuan Li, Haotian Liu, Liunian Li, Pengchuan Zhang, Jyoti Aneja, Jianwei Yang, Ping Jin, Houdong Hu, Zicheng Liu, Yong~Jae Lee, et~al.
\newblock Elevater: A benchmark and toolkit for evaluating language-augmented visual models.
\newblock \emph{Advances in Neural Information Processing Systems}, 35:\penalty0 9287--9301, 2022{\natexlab{a}}.

\bibitem[Li et~al.(2022{\natexlab{b}})Li, Li, Xiong, and Hoi]{li2022blip}
Junnan Li, Dongxu Li, Caiming Xiong, and Steven Hoi.
\newblock Blip: Bootstrapping language-image pre-training for unified vision-language understanding and generation.
\newblock In \emph{International conference on machine learning}, pages 12888--12900. PMLR, 2022{\natexlab{b}}.

\bibitem[Li et~al.(2023)Li, Li, Savarese, and Hoi]{li2023blip}
Junnan Li, Dongxu Li, Silvio Savarese, and Steven Hoi.
\newblock Blip-2: Bootstrapping language-image pre-training with frozen image encoders and large language models.
\newblock In \emph{International conference on machine learning}, pages 19730--19742. PMLR, 2023.

\bibitem[Lin et~al.(2024)Lin, Yin, Ping, Molchanov, Shoeybi, and Han]{lin2024vila}
Ji Lin, Hongxu Yin, Wei Ping, Pavlo Molchanov, Mohammad Shoeybi, and Song Han.
\newblock Vila: On pre-training for visual language models.
\newblock In \emph{Proceedings of the IEEE/CVF Conference on Computer Vision and Pattern Recognition}, pages 26689--26699, 2024.

\bibitem[Liu et~al.(2024)Liu, Li, Wu, and Lee]{liu2024visual}
Haotian Liu, Chunyuan Li, Qingyang Wu, and Yong~Jae Lee.
\newblock Visual instruction tuning.
\newblock \emph{Advances in neural information processing systems}, 36, 2024.

\bibitem[McKinzie et~al.(2024)McKinzie, Gan, Fauconnier, Dodge, Zhang, Dufter, Shah, Du, Peng, Weers, et~al.]{mckinzie2024mm1}
Brandon McKinzie, Zhe Gan, Jean-Philippe Fauconnier, Sam Dodge, Bowen Zhang, Philipp Dufter, Dhruti Shah, Xianzhi Du, Futang Peng, Floris Weers, et~al.
\newblock Mm1: Methods, analysis \& insights from multimodal llm pre-training.
\newblock \emph{arXiv preprint arXiv:2403.09611}, 2024.

\bibitem[Peng et~al.(2024)Peng, Xie, You, Lan, and Wu]{peng2024synthesize}
Wujian Peng, Sicheng Xie, Zuyao You, Shiyi Lan, and Zuxuan Wu.
\newblock Synthesize diagnose and optimize: Towards fine-grained vision-language understanding.
\newblock In \emph{Proceedings of the IEEE/CVF Conference on Computer Vision and Pattern Recognition}, pages 13279--13288, 2024.

\bibitem[Podell et~al.(2023)Podell, English, Lacey, Blattmann, Dockhorn, M{\"u}ller, Penna, and Rombach]{podell2023sdxl}
Dustin Podell, Zion English, Kyle Lacey, Andreas Blattmann, Tim Dockhorn, Jonas M{\"u}ller, Joe Penna, and Robin Rombach.
\newblock Sdxl: Improving latent diffusion models for high-resolution image synthesis.
\newblock \emph{arXiv preprint arXiv:2307.01952}, 2023.

\bibitem[Radford et~al.(2021)Radford, Kim, Hallacy, Ramesh, Goh, Agarwal, Sastry, Askell, Mishkin, Clark, et~al.]{radford2021learning}
Alec Radford, Jong~Wook Kim, Chris Hallacy, Aditya Ramesh, Gabriel Goh, Sandhini Agarwal, Girish Sastry, Amanda Askell, Pamela Mishkin, Jack Clark, et~al.
\newblock Learning transferable visual models from natural language supervision.
\newblock In \emph{International conference on machine learning}, pages 8748--8763. PMLR, 2021.

\bibitem[Schuhmann et~al.(2021)Schuhmann, Vencu, Beaumont, Kaczmarczyk, Mullis, Katta, Coombes, Jitsev, and Komatsuzaki]{schuhmann2021laion}
Christoph Schuhmann, Richard Vencu, Romain Beaumont, Robert Kaczmarczyk, Clayton Mullis, Aarush Katta, Theo Coombes, Jenia Jitsev, and Aran Komatsuzaki.
\newblock Laion-400m: Open dataset of clip-filtered 400 million image-text pairs.
\newblock \emph{arXiv preprint arXiv:2111.02114}, 2021.

\bibitem[Singh et~al.(2022)Singh, Hu, Goswami, Couairon, Galuba, Rohrbach, and Kiela]{singh2022flava}
Amanpreet Singh, Ronghang Hu, Vedanuj Goswami, Guillaume Couairon, Wojciech Galuba, Marcus Rohrbach, and Douwe Kiela.
\newblock Flava: A foundational language and vision alignment model.
\newblock In \emph{Proceedings of the IEEE/CVF Conference on Computer Vision and Pattern Recognition}, pages 15638--15650, 2022.

\bibitem[Thomee et~al.(2016)Thomee, Shamma, Friedland, Elizalde, Ni, Poland, Borth, and Li]{thomee2016yfcc100m}
Bart Thomee, David~A Shamma, Gerald Friedland, Benjamin Elizalde, Karl Ni, Douglas Poland, Damian Borth, and Li-Jia Li.
\newblock Yfcc100m: The new data in multimedia research.
\newblock \emph{Communications of the ACM}, 59\penalty0 (2):\penalty0 64--73, 2016.

\bibitem[Thrush et~al.(2022)Thrush, Jiang, Bartolo, Singh, Williams, Kiela, and Ross]{thrush2022winoground}
Tristan Thrush, Ryan Jiang, Max Bartolo, Amanpreet Singh, Adina Williams, Douwe Kiela, and Candace Ross.
\newblock Winoground: Probing vision and language models for visio-linguistic compositionality.
\newblock In \emph{Proceedings of the IEEE/CVF Conference on Computer Vision and Pattern Recognition}, pages 5238--5248, 2022.

\bibitem[Urbanek et~al.(2024)Urbanek, Bordes, Astolfi, Williamson, Sharma, and Romero-Soriano]{urbanek2024picture}
Jack Urbanek, Florian Bordes, Pietro Astolfi, Mary Williamson, Vasu Sharma, and Adriana Romero-Soriano.
\newblock A picture is worth more than 77 text tokens: Evaluating clip-style models on dense captions.
\newblock In \emph{Proceedings of the IEEE/CVF Conference on Computer Vision and Pattern Recognition}, pages 26700--26709, 2024.

\bibitem[Xue et~al.(2024)Xue, Shu, Awadalla, Wang, Yan, Purushwalkam, Zhou, Prabhu, Dai, Ryoo, et~al.]{xue2024xgen}
Le Xue, Manli Shu, Anas Awadalla, Jun Wang, An Yan, Senthil Purushwalkam, Honglu Zhou, Viraj Prabhu, Yutong Dai, Michael~S Ryoo, et~al.
\newblock xgen-mm (blip-3): A family of open large multimodal models.
\newblock \emph{arXiv preprint arXiv:2408.08872}, 2024.

\bibitem[Yarom et~al.(2024)Yarom, Bitton, Changpinyo, Aharoni, Herzig, Lang, Ofek, and Szpektor]{yarom2024you}
Michal Yarom, Yonatan Bitton, Soravit Changpinyo, Roee Aharoni, Jonathan Herzig, Oran Lang, Eran Ofek, and Idan Szpektor.
\newblock What you see is what you read? improving text-image alignment evaluation.
\newblock \emph{Advances in Neural Information Processing Systems}, 36, 2024.

\bibitem[Yuksekgonul et~al.(2023)Yuksekgonul, Bianchi, Kalluri, Jurafsky, and Zou]{yuksekgonul2023and}
Mert Yuksekgonul, Federico Bianchi, Pratyusha Kalluri, Dan Jurafsky, and James Zou.
\newblock When and why vision-language models behave like bags-of-words, and what to do about it?
\newblock In \emph{The Eleventh International Conference on Learning Representations}, 2023.

\bibitem[Zeng et~al.(2021)Zeng, Zhang, and Li]{zeng2021multi}
Yan Zeng, Xinsong Zhang, and Hang Li.
\newblock Multi-grained vision language pre-training: Aligning texts with visual concepts.
\newblock \emph{arXiv preprint arXiv:2111.08276}, 2021.

\bibitem[Zeng et~al.(2024)Zeng, Huang, Zhang, Jie, Chai, and Wang]{zeng2024investigating}
Yunan Zeng, Yan Huang, Jinjin Zhang, Zequn Jie, Zhenhua Chai, and Liang Wang.
\newblock Investigating compositional challenges in vision-language models for visual grounding.
\newblock In \emph{Proceedings of the IEEE/CVF Conference on Computer Vision and Pattern Recognition}, pages 14141--14151, 2024.

\bibitem[Zhai et~al.(2023)Zhai, Mustafa, Kolesnikov, and Beyer]{zhai2023sigmoid}
Xiaohua Zhai, Basil Mustafa, Alexander Kolesnikov, and Lucas Beyer.
\newblock Sigmoid loss for language image pre-training.
\newblock In \emph{Proceedings of the IEEE/CVF International Conference on Computer Vision}, pages 11975--11986, 2023.

\bibitem[Zhao et~al.(2022)Zhao, Zhang, Zhu, Shen, Lee, Lu, and Yin]{zhao2022vl}
Tiancheng Zhao, Tianqi Zhang, Mingwei Zhu, Haozhan Shen, Kyusong Lee, Xiaopeng Lu, and Jianwei Yin.
\newblock Vl-checklist: Evaluating pre-trained vision-language models with objects, attributes and relations.
\newblock \emph{arXiv preprint arXiv:2207.00221}, 2022.

\bibitem[Zhu et~al.(2023)Zhu, Chen, Shen, Li, and Elhoseiny]{zhu2023minigpt}
Deyao Zhu, Jun Chen, Xiaoqian Shen, Xiang Li, and Mohamed Elhoseiny.
\newblock Minigpt-4: Enhancing vision-language understanding with advanced large language models.
\newblock \emph{arXiv preprint arXiv:2304.10592}, 2023.

\end{thebibliography}
}

\end{document}